\pdfoutput=1
\documentclass{article}

\PassOptionsToPackage{numbers,compress}{natbib}

\usepackage[main, final]{neurips_2026}

\usepackage[utf8]{inputenc} %
\usepackage[T1]{fontenc}    %
\usepackage{hyperref}       %
\usepackage{url}            %
\usepackage{booktabs}       %
\usepackage{amsfonts}       %
\usepackage{amsmath}        %
\usepackage{mathtools}      %
\usepackage{mdframed}       %
\usepackage[most]{tcolorbox} %
\usepackage{subcaption}     %
\usepackage{tikz}           %
\usetikzlibrary{arrows.meta}
\usepackage{nicefrac}       %
\usepackage{microtype}      %
\usepackage{xcolor}         %
\usepackage{pgf}            %
\usepackage{fontawesome5}   %
\definecolor{pagelink}{HTML}{1E3F99}  %

\title{The Shape of a Program: Path Signatures for Trace-to-Program Induction}

\author{Mohamed Ghanem$^{1}$ \quad Bernd Finkbeiner$^{1, 2}$ \\
$^1$CISPA Helmholtz Center for Information Security \\
$^2$Technical University of Munich \\
\texttt{mohamed.ghanem@cispa.de}, \texttt{finkbeiner@cispa.de}
}

\begin{document}

\maketitle

\begin{abstract}
Synthesizing programs from execution traces is a fundamental challenge in algorithmic interpretability and the reverse engineering of complex systems. We bring a rough-path-theoretic perspective to program induction, treating execution traces as high-dimensional rough paths and encoding them with the path signature transform, a tool from stochastic analysis whose expectation provably characterizes the distribution of traces a program induces. On three program-induction domains, Karel, List, and RASP, our log-signature trace encoder outperforms LSTM, Transformer, and Fourier baselines in almost every setting, with its largest margins when training data is scarce. We further observe that the trace-to-program pipeline is naturally contractive: it often produces valid programs that are shorter than the ground truth. We turn this into a bootstrapping procedure that exposes redundancy in the ground-truth supervision without any external oracle, which we demonstrate on redundancy-injected versions of our datasets. Finally, the log-signature encoder induces a markedly more interpretable latent geometry than the baselines, in which the nearest neighbors of every state lie on its own execution trace.

\end{abstract}
\vspace{-0.5em}
\begin{center}
  {\hypersetup{pdfborder={0 0 0}}\color{pagelink}\faGlobe~\url{https://oschart.github.io/program-signature}}
\end{center}

\section{Introduction}

Program synthesis, the task of automatically generating code that satisfies a given specification, stands as one of the central challenges in modern artificial intelligence. While much of the recent progress has focused on synthesizing programs from functional specifications, typically provided as input-output (I/O) pairs \citep{devlin2017robustfill,balog2016deepcoder,bunel2018leveraging,chen2018execution}, this represents only one facet of the problem. A distinct but equally important challenge is synthesizing programs from behavioral specifications in the form of execution traces. This task is commonly referred to as \textit{program induction}. Unlike I/O pairs, which treat the program as a black box, execution traces reveal the how of the computation, offering a granular view of the algorithmic decision-making process. Mastering this form of synthesis is essential for real-world applications such as reverse-engineering legacy systems, malware analysis, and algorithmic interpretability, where the goal is not just to replicate an output, but to recover the underlying logic that produced a specific sequence of states.

This direction is essential for algorithmic interpretability, as it allows us to \textit{distill} opaque agents into human-readable code \citep{verma2018programmatically}. By synthesizing programs that faithfully reproduce the trace of a policy rather than just its final output, we gain verifiable insights into agent behavior, revealing safety constraints and logic gaps that are otherwise hidden in high-dimensional weight matrices. Towards this end, it is compelling to find more optimal methodologies for encoding behavioral traces. Unlike static I/O pairs, behavioral traces possess intrinsic temporal and geometric structure that characterizes how a task is solved, not just what the solution is. Consequently, the success of the synthesis process hinges on the representation's ability to capture these dynamics robustly.

To tackle this task, the dominant paradigm has been to treat execution traces as discrete sequences, processing them with standard sequence models like LSTMs \citep{hochreiter1997long} or Transformers \citep{vaswani2017attention}. In this work, we propose a fundamental shift in perspective. Rather than viewing traces merely as sequences of tokens, we conceptualize them as high-dimensional rough paths and encode them using the \textit{path signature transform}, a tool from stochastic analysis and rough path theory \citep{chen1957integration} that has been of growing interest in machine learning for time-series forecasting and reinforcement learning \citep{chevyrev2025primer,kidger2019deep}. Under this lens, a program is a generator of rough paths: each execution traces out a curve in the latent state space whose geometric shape reflects the underlying control flow. The signature transform turns that geometry into a fixed-size, length-invariant embedding with a principled algebraic structure, providing a representation that is naturally aligned with how programs structure execution.

Several properties of the signature reinforce this perspective. \textit{First, the expected signature characterizes the trace distribution a program induces} \citep{chevyrev2016characteristic}, so aggregating signatures across executions of the same program can, in principle, identify it. \textit{Second, signatures factor under any temporal decomposition of the path}: by Chen's identity, $S(X * Y) = S(X) \otimes S(Y)$, so the signature of a trace is the product of the signatures of its parts (Section~\ref{subsec:why_signatures}). \textit{Third}, the signature captures the shape of the path in a fixed-size representation regardless of path length. \textit{Fourth}, its runtime is linear in the path length, making it much more practical for processing long traces than Transformers with their quadratic attention complexity (Appendix~\ref{subsec:wall_clock}). Beyond these four points, the signature is a \textit{universal nonlinearity}: any continuous functional of the path can be linearly approximated from its signature, so no expressive power is sacrificed for these structural advantages. This last property notably does not hold for classical operators like the Fourier transform.

\paragraph{Contributions.} We make the following key contributions:
{\settowidth{\leftmargini}{(iii)\hspace{\labelsep}}%
\begin{enumerate}
    \item[(i)] We explore the use of path signature methods for learning programmatic behavior, bringing a rough-path-theoretic perspective into program induction. On three program-induction domains, our log-signature trace encoder outperforms LSTM, Transformer, and Fourier baselines in almost every setting, with its largest margins when training data is scarce.
    \item[(ii)] We show that our trace-to-program pipeline is naturally \emph{contractive}: it produces shorter valid programs than the ground-truth program in a sizable fraction of cases. We turn this property into a \emph{bootstrapping} procedure that progressively replaces ground-truth training programs with model-generated ones whenever they are valid and shorter, and use it as a measurement instrument that exposes redundancy in the supervision without any external oracle.
    \item[(iii)] We show that the log-signature encoder induces a markedly more interpretable latent geometry than the baselines: the states of each execution trace form a separate chain, and the nearest neighbors of every state lie on its own trace.
\end{enumerate}}

\section{Related work}

\paragraph{Path signature in deep learning.}
Early applications utilized the truncated signature as a static feature vector for standard supervised learning tasks, such as handwriting analysis and financial time-series classification \citep{lai2019offline,gyurko2013extracting}. It has also been used to summarize trajectories in non-Markovian reinforcement learning as an agent memory \citep{kidger2019deep}. More recently, Neural Rough Differential Equations \citep{morrill2021neural} have used the log-signature transform as the driving control of a continuous-time architecture for long time series, embedding signature theory directly into the model's dynamics rather than treating it as a fixed feature map. A common thread between these methods is that they aim to \textit{extract} some information from the signature features. In contrast, we use the signature of a behavioral trace not only to summarize it but also to characterize and reconstruct the generating process, i.e., the program.

\paragraph{Neural program induction.}
This paradigm leverages the insight that traces provide a denser supervisory signal than I/O pairs, reducing the search space for complex algorithms. Many existing works learn the programs in the latent weights of the model \citep{neelakantan2015neural, graves2014neural,yan2020neural}. This has the downside of losing interpretability over the learned programs. \citet{reed2015neural} introduced the Neural Programmer-Interpreter (NPI), which learned to execute programs by composing sub-programs, trained to imitate full execution traces, though it required expensive step-by-step supervision. To relax this requirement, \citet{shin2018improving} proposed a two-stage trace-inference framework, where a model first predicts a latent execution trace from I/O pairs and subsequently decodes the program code from that predicted trace, significantly improving performance on DSL tasks like Karel. Closest to our setting, \citet{sun2018neural} synthesize Karel programs from multiple demonstrations, i.e., sequences of environment states, which they encode with recurrent networks. In this paper, we take a simpler and more focused approach that employs an encoder-decoder architecture with the ultimate goal of evaluating the efficacy of different trace encoders.

\section{Preliminaries}

Let $X: [0, T] \to \mathbb{R}^d$ be a path. The signature of $X$, denoted by $S(X)$, is an infinite graded sequence of iterated integrals that provides a faithful representation of the path's geometric shape.

Formally, the signature is an element of the extended tensor algebra $T((\mathbb{R}^d)) = \prod_{\ell=0}^\infty (\mathbb{R}^d)^{\otimes \ell}$. It is defined as the collection of all iterated integrals of the path coordinates:
\begin{equation}
    S(X)_{0,T} = \left( 1, \mathbf{X}^{(1)}_{0,T}, \mathbf{X}^{(2)}_{0,T}, \dots, \mathbf{X}^{(\ell)}_{0,T}, \dots \right),
\end{equation}
where the $\ell$-th level term $\mathbf{X}^{(\ell)}$ is a tensor in $(\mathbb{R}^d)^{\otimes \ell}$ with components given by:
\begin{equation}
    \mathbf{X}^{(\ell)}_{0,T} = \int_{0 < t_1 < \dots < t_\ell < T} dX_{t_1} \otimes \dots \otimes dX_{t_\ell}.
\end{equation}
The term at level $\ell=0$ is conventionally $1$.

In the context of discrete data analysis, we typically observe a discrete time series $\mathbf{x} = (x_1, x_2, \dots, x_N)$ where $x_i \in \mathbb{R}^d$. To compute the signature, we treat $\mathbf{x}$ as a continuous path by applying a piecewise linear interpolation. Let $X$ be the continuous path obtained by linearly interpolating between the points in $\mathbf{x}$. 

Utilizing \textit{Chen's identity} \citep{chen1957integration}, which states that $S(X * Y) = S(X) \otimes S(Y)$ where $X * Y$ denotes their concatenation, the signature of the discrete path can be computed efficiently as the tensor product of the signatures of its linear segments:
\begin{equation}
    S(\mathbf{x}) = \bigotimes_{i=1}^{N-1} \exp\left( x_{i+1} - x_i \right),
\end{equation}
where the exponential map is defined within the tensor algebra.

\paragraph{Key properties and invariances.}
The path signature possesses several algebraic and geometric properties \citep{chevyrev2025primer} that make it a highly desirable feature map for machine learning tasks.
Firstly, the signature map is injective for paths up to tree-like equivalence \citep{hambly2010uniqueness}. Essentially, $S(X)$ uniquely determines the path $X$ (up to ``backtracking''), ensuring that no information regarding the shape of the trajectory is lost during the transformation.
Secondly, the signature is a \textit{universal nonlinearity}: linear functionals on the signature can approximate any continuous function of the path arbitrarily well. This allows linear models (like linear regression) trained on signature features to learn complex, nonlinear relationships in the original data space.

\paragraph{Truncation and log-signature.}
The full signature is an infinite-dimensional object. For computational feasibility, we compute the \textbf{truncated signature} up to level $M$, denoted as $S^M(X)$. The dimension of the feature vector for a path in $\mathbb{R}^d$ truncated at level $M$ is $\sum_{\ell=1}^M d^\ell$, which grows geometrically with $M$.

To mitigate this dimensionality explosion, we utilize the \textbf{log-signature} $\log(S(X))$. The logarithm of the signature maps the elements from the tensor algebra into the underlying Lie algebra. The log-signature is a more compact representation that retains the same information as the signature (up to tree-like equivalence) but removes the redundant terms generated by the shuffle product.

The dimension of the log-signature at level $\ell$ is given by Witt's formula, which is significantly smaller than $d^\ell$. For example, while the signature of a path in $\mathbb{R}^2$ truncated at $M=4$ has 30 terms, the log-signature has only 8.

\providecolor{archline}{HTML}{666666}        %
\providecolor{archfill}{HTML}{F0F0F0}        %
\providecolor{archcard}{HTML}{4D4D4D}        %
\providecolor{archwall}{HTML}{808080}
\providecolor{archmarker}{HTML}{E3C800}
\providecolor{archmarkerdraw}{HTML}{B09500}
\providecolor{archaccent}{HTML}{2F5BD3}      %
\providecolor{archaccentdark}{HTML}{1E3F99}
\providecolor{archaccentlight}{HTML}{D7E3FB}
\providecolor{archaccenttext}{HTML}{17307A}
\providecolor{archband}{HTML}{EEF3FD}
\providecolor{archbanddraw}{HTML}{AFC3EE}
\providecolor{archpos}{HTML}{C1CEF2}         %
\providecolor{archneg}{HTML}{F4D1BA}         %
\providecolor{archkw}{HTML}{8B0000}          %
\providecolor{archtok}{HTML}{6B6B6B}
\providecolor{archhl}{HTML}{FFE9A8}          %
\newlength{\archunit}
\tikzset{
  arch tip/.style={-{Stealth[length=2.2pt, width=1.8pt]}},
  arch train/.style={arch tip, black!55, dash pattern=on 1.5pt off 1pt},  %
  arch box/.style={draw=archline, fill=#1, rounded corners=2pt, inner sep=0pt, font=\fontsize{6.5}{7}\selectfont},
  arch key/.pic={
    \filldraw[fill=archmarker, draw=archmarkerdraw, line width=0.1pt]
      (0,-3.2) circle[radius=2.2] (-0.8,-1.2) rectangle (0.8,5.5) (0.8,3) rectangle (2.2,4.2);
  },
  arch agent/.pic={%
    \fill (-2.8,-8.4) -- (2.8,-8.4) -- (2.8,0) -- (7.4,0) -- (0,8.4) -- (-7.4,0) -- (-2.8,0) -- cycle;
  },
}
\providecommand{\archgrid}{%
  \fill[archwall] (49.2,0) rectangle ++(16.4,16.4) (16.4,32.8) rectangle ++(16.4,32.8);
  \foreach \i in {0,...,4} \draw[line width=0.3pt] (\i*16.4,0) -- ++(0,65.6) (0,\i*16.4) -- ++(65.6,0);
}
\providecommand{\archcell}[2]{({(#1+0.5)*16.4},{(#2+0.5)*16.4})}
\providecommand{\archprog}[1]{%
  \textcolor{archkw}{def run}():\\
  \hspace*{0.8em}\textcolor{archkw}{while not} \textcolor{archtok}{markersPresent}():\\
  \hspace*{1.6em}\textcolor{archtok}{move}()\\
  \hspace*{0.8em}{\setlength{\fboxsep}{0.4pt}\colorbox{archhl}{\strut\textcolor{archtok}{#1}}}()}

\begin{figure*}[h]
  \begin{center}
  \setlength{\archunit}{\dimexpr 0.7\columnwidth/838\relax}
  \begin{tikzpicture}[x=\archunit, y=-\archunit, line width=0.4pt]
    \foreach \o in {8,4} \filldraw[fill=white, draw=black!45, rounded corners=3pt] (231+\o,347-\o) rectangle (750+\o,440-\o);
    \filldraw[fill=white, rounded corners=3pt] (231,347) rectangle (750,440);
    \node[anchor=north west, inner sep=2pt, font=\fontsize{5.5}{5.5}\selectfont] at (231,347) {\texttimes\textit{K}};
    \node[draw=archcard, rounded corners=3pt, minimum width=174.7\archunit, minimum height=81.4\archunit] (rho) at (7.35,393.5) {};
    \node[anchor=west, align=left, inner sep=0pt, font=\fontsize{4.5}{5.6}\selectfont] at (-76,393.5) {\archprog{pickMarker}};
    \node[fill=white, inner sep=1pt, text=archcard, font=\fontsize{7}{7}\selectfont] at (78,352.8) {$\rho$};
    \draw[arch tip] (rho.east) -- node[above, text=archcard, font=\fontsize{6}{7}\ttfamily] {exec} (231,393.5);

    \begin{scope}[shift={(267.2,360.7)}]
      \archgrid \pic at \archcell{1}{0} {arch key}; \pic at \archcell{2}{2} {arch key}; \pic at \archcell{2}{0} {arch agent};
    \end{scope}
    \begin{scope}[shift={(395.5,360.7)}]
      \archgrid \pic at \archcell{1}{0} {arch key}; \pic at \archcell{2}{2} {arch key}; \pic at \archcell{2}{1} {arch agent};
    \end{scope}
    \begin{scope}[shift={(518.7,360.7)}]
      \archgrid \pic at \archcell{1}{0} {arch key}; \pic at \archcell{2.25}{1.8} {arch key};
      \pic[scale=0.7] at \archcell{1.85}{2.15} {arch agent};
    \end{scope}
    \begin{scope}[shift={(649.5,360.7)}]
      \archgrid \pic at \archcell{1}{0} {arch key}; \pic at \archcell{2}{2} {arch agent};
    \end{scope}
    \begin{scope}[font=\fontsize{3.8}{3.8}\selectfont, every node/.style={above, inner sep=1.5pt}]
      \draw[arch tip] (332.8,393.5) -- node {\textcolor{archtok}{move}()} (395.5,393.5);
      \draw[arch tip] (461.1,393.5) -- node {\textcolor{archtok}{move}()} (518.7,393.5);
      \draw[arch tip] (584.3,393.5) -- node {\textcolor{archtok}{pickMarker}()} (649.5,393.5);
    \end{scope}

    \filldraw[fill=black!3, draw=black!40, dash pattern=on 2pt off 1.5pt, rounded corners=4pt] (207,500) rectangle (756,728);
    \node[anchor=south east, align=right, inner sep=0pt, text=black!60, font=\fontsize{5.5}{6.2}\selectfont\itshape]
      at (750,723) {Log-signature\\trace encoder};
    \filldraw[fill=archband, draw=archbanddraw, rounded corners=3pt] (237.5,580) rectangle (745.2,635.7);
    \node[anchor=north west, inner sep=2pt, text=archaccent, font=\fontsize{7}{7}\selectfont] at (237.5,580) {$\mathbb{R}^d$};
    \foreach \x/\py [count=\t from 0] in {300/623.5, 428.3/590.7, 551.5/626.8, 682.3/595.7} \coordinate (p\t) at (\x,\py);
    \fill[archneg] (p0) -- (p1) -- (492.5,609.5) -- cycle;
    \fill[archpos] (492.5,609.5) -- (p2) -- (p3) -- cycle;
    \draw[arch tip, black!75, dash pattern=on 1.5pt off 1pt, shorten >=1.5pt] (p0) -- (p3);
    \node[font=\fontsize{5}{5}\selectfont] at (406.9,607.9) {$-$};
    \node[font=\fontsize{5}{5}\selectfont] at (575.4,610.7) {$+$};
    \foreach \x [count=\t from 0] in {300, 428.3, 551.5, 682.3} {
      \draw[arch tip, dash pattern=on 0.4pt off 0.8pt] (\x,426.3) -- (\x,449);
      \node[arch box=archfill, minimum width=50.6\archunit, minimum height=42.3\archunit, font=\fontsize{9}{9}\selectfont] at (\x,470.2) {$\phi$};
      \draw[arch tip] (\x,491.3) -- (\x,519.7);
      \draw[fill=white] (\x,530) circle[radius=10.3] (\x-10.3,530) -- (\x+10.3,530) (\x,519.7) -- (\x,540.3);
      \node[arch box=white, minimum width=50.6\archunit, minimum height=42.3\archunit, font=\fontsize{5.5}{5.5}\selectfont]
        (pe\t) at (\x-58,530) {PE(\t)};
      \draw[arch tip] (pe\t.east) -- (\x-10.3,530);
      \draw[arch tip, black!70, line width=0.6pt, dash pattern=on 0.6pt off 1.2pt] (\x,540.3) -- (p\t);
    }
    \foreach \a/\b in {0/1, 1/2, 2/3} \draw[arch tip, archaccent, line width=0.7pt, shorten >=1pt] (p\a) -- (p\b);
    \foreach \t in {0,...,3} \fill[archaccent] (p\t) circle[radius=3];
    \node[draw=archaccentdark, fill=archaccent, text=white, rounded corners=2pt, minimum width=186.6\archunit,
      minimum height=48.8\archunit, font=\fontsize{6.5}{7}\selectfont\bfseries] (logsig) at (488,693.4) {Log-signature};
    \foreach \t in {0,...,3} \draw[black!45, line width=0.3pt, dash pattern=on 1pt off 1pt] (p\t) -- (logsig.north);
    \node[draw=archaccent, fill=archaccentlight, text=archaccenttext, rounded corners=2pt, minimum width=147.3\archunit,
      minimum height=30.9\archunit, font=\fontsize{6.5}{7}\selectfont] (proj) at (295.6,693.4) {Projection};
    \draw[arch tip] (logsig.west) -- (proj.east);

    \filldraw[fill=archfill, draw=archline, rounded corners=2pt]
      (131.3,614.1) -- (175.2,634.1) -- (175.2,752.9) -- (131.3,772.9) -- cycle;
    \node[rotate=-90, font=\fontsize{6.5}{7}\selectfont] at (153.2,693.5) {Decoder};
    \draw[arch tip] (proj.west) -- (175.2,693.4);
    \node[above, inner sep=1pt, font=\fontsize{6}{6}\selectfont] at (191.1,693.4) {$\mathbf{h}_k$};
    \node[draw=archcard, rounded corners=3pt, minimum width=174.7\archunit, minimum height=81.4\archunit] (prog) at (7.35,693.4) {};
    \node[anchor=west, align=left, inner sep=0pt, font=\fontsize{4.5}{5.6}\selectfont] at (-76,693.4) {\archprog{putMarker}};
    \node[fill=white, inner sep=1pt, text=archcard, font=\fontsize{7}{7}\selectfont] at (78,652.7) {$\hat\rho$};
    \draw[arch tip] (131.3,693.4) -- (prog.east);

    \node[arch box=white, minimum width=130\archunit, minimum height=45\archunit] (loss) at (7.35,542.65) {Cross-entropy};
    \begin{scope}[every node/.style={left, inner sep=2pt, text=black!60, font=\fontsize{5}{5}\selectfont\itshape}]
      \draw[arch train] (rho.south) -- node {target} (loss.north);
      \draw[arch train] (prog.north) -- node {prediction} (loss.south);
    \end{scope}
  \end{tikzpicture}
  \caption{
    Overall trace-to-program architecture. 
  }
  \label{fig:main}
  \end{center}
\end{figure*}
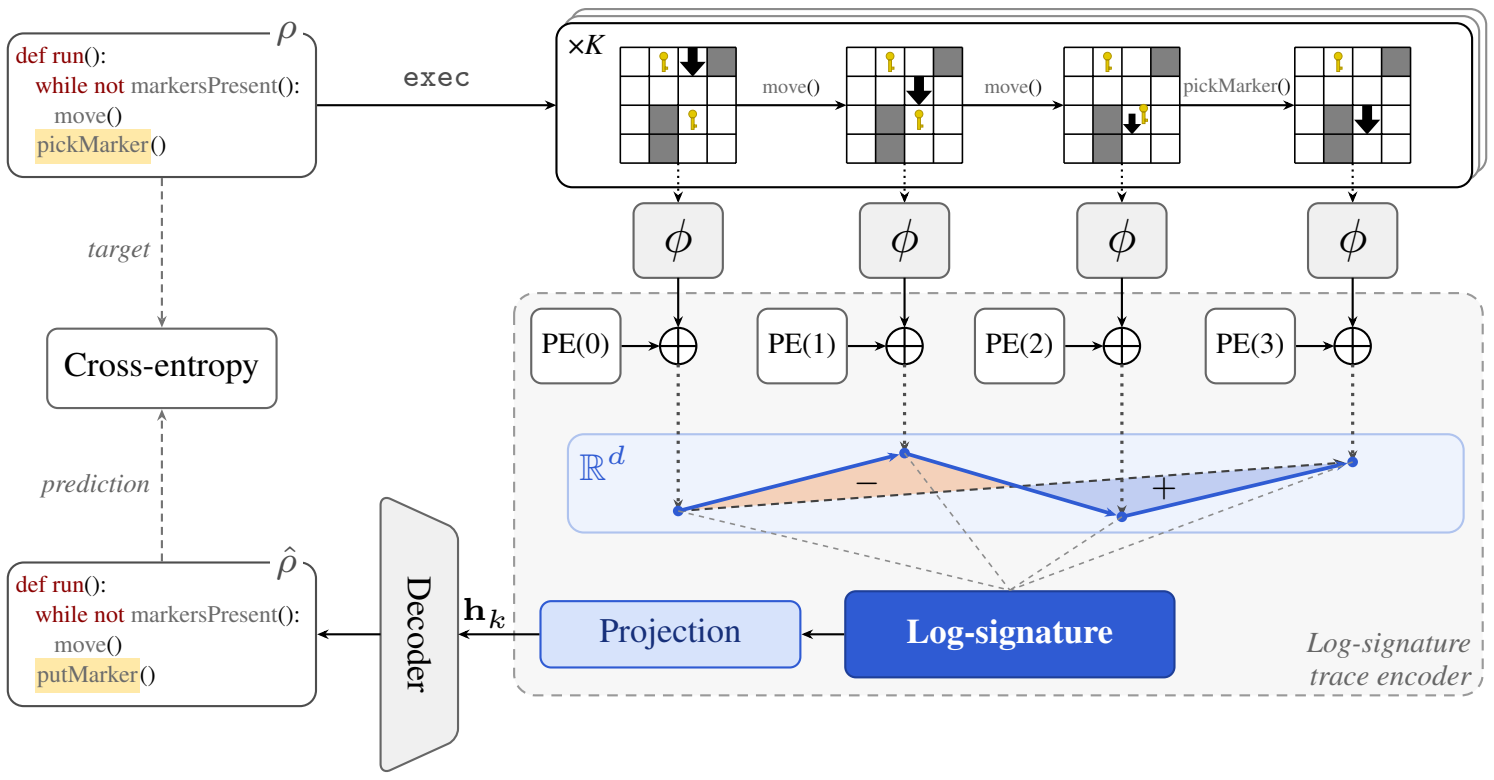

\section{Approach}
We propose an encoder-decoder architecture for synthesizing programs from execution traces. Our pipeline consists of three key networks: (1) a state encoder $\phi$ that embeds the semantic state of the environment, where each state embedding is a $d$-dimensional point on the latent trace. Next, we use (2) a trace encoder that embeds state sequences into a single representation that gets passed to (3) a Transformer-based program decoder to synthesize the trace-generating program. Our trace encoder is based on the path log-signature transform, which captures the geometric properties of the state sequence. An overview of our architecture is shown in Figure~\ref{fig:main}. Note that Karel grids are just one example of a domain to which our approach applies; the same architecture and training procedure are used for the List and RASP domains.

\subsection{Problem formulation}
We consider the general task of \emph{trace-to-program synthesis}: given a set of execution traces produced by an unknown program, recover a program in a target domain-specific language (DSL) that reproduces them. Formally, fix a DSL $\mathcal{P}$ and a runtime environment $\mathcal{E}$ that endows $\mathcal{P}$ with a state space $\mathcal{S}$ and an execution semantics: running a program $\rho \in \mathcal{P}$ from an initial state $s_0 \in \mathcal{S}$ produces a trace
\begin{equation}
    \tau = \mathcal{E}(\rho, s_0) = (s_1, s_2, \dots, s_T), \qquad s_t \in \mathcal{S},
\end{equation}
where $T$ is the (program- and state-dependent) execution length. Given a set of traces $\mathcal{T} = \{\tau_1, \dots, \tau_K\}$ produced by a hidden program $\rho^*$ from initial states $\{s_{k,0}\}_{k=1}^K$, the goal is to learn a mapping $\mathcal{T} \mapsto \hat\rho$ such that $\mathcal{E}(\hat\rho, s_{k,0}) = \tau_k$ for every $k$. Different choices of $(\mathcal{P}, \mathcal{E})$ instantiate different program-induction domains.
Throughout, we use $\tau_k = (s_{k,1}, \dots, s_{k,T_k})$ to denote the $k$-th trace and $T_k$ its length.

\providecolor{dslnt}{HTML}{1F4E79}
\providecolor{dsltm}{HTML}{6B6B6B}
\providecolor{dslkw}{HTML}{8B0000}
\providecommand{\dslnt}[1]{\textcolor{dslnt}{\textit{#1}}}
\providecommand{\dslkw}[1]{\textcolor{dslkw}{\texttt{#1}}}
\providecommand{\dsltm}[1]{\textcolor{dsltm}{\texttt{#1}}}
\providecommand{\dslbnf}{\ensuremath{\;::=\;}}
\providecommand{\dslalt}{\ensuremath{\;\mid\;}}
\providecommand{\dslfont}{\fontsize{6}{7.2}\selectfont}

\begin{figure}[ht]
\centering
\begin{subfigure}[t]{0.333\linewidth}
  \centering
  \begin{tcolorbox}[
      enhanced, sharp corners, boxrule=0.5pt, colframe=black, colback=white,
      left=2pt, right=2pt, top=3pt, bottom=3pt,
      equal height group=dslboxes,
  ]
  \dslfont
  \setlength{\tabcolsep}{1pt}
  \renewcommand{\arraystretch}{1.1}
  \begin{tabular}{r@{\;}c@{\;}l}
  \dslnt{Prog}    & \dslbnf & \dslkw{DEF}\,\dslkw{run}\,\dsltm{m(}\;\dslnt{Stmt}\;\dsltm{m)} \\[1pt]
  \dslnt{Stmt}    & \dslbnf & \dslnt{Action} \dslalt \dslnt{Stmt}\;\dslnt{Stmt} \\
                  &         & \dslalt\;\dslnt{If} \dslalt \dslnt{IfElse} \dslalt \dslnt{While} \dslalt \dslnt{Repeat} \\[1pt]
  \dslnt{If}      & \dslbnf & \dslkw{IF}\;\dsltm{c(}\,\dslnt{Cond}\,\dsltm{c)}\;\dsltm{i(}\,\dslnt{Stmt}\,\dsltm{i)} \\
  \dslnt{IfElse}  & \dslbnf & \dslkw{IFELSE}\;\dsltm{c(}\,\dslnt{Cond}\,\dsltm{c)} \\
                  &         & \dsltm{i(}\,\dslnt{Stmt}\,\dsltm{i)}\;\dslkw{ELSE}\;\dsltm{e(}\,\dslnt{Stmt}\,\dsltm{e)} \\
  \dslnt{While}   & \dslbnf & \dslkw{WHILE}\;\dsltm{c(}\,\dslnt{Cond}\,\dsltm{c)}\;\dsltm{w(}\,\dslnt{Stmt}\,\dsltm{w)} \\
  \dslnt{Repeat}  & \dslbnf & \dslkw{REPEAT}\;\dslnt{N}\;\dsltm{r(}\,\dslnt{Stmt}\,\dsltm{r)} \\
                  &         & \quad{\tiny($\dslnt{N}\!\in\![0,19]$)} \\[1pt]
  \dslnt{Cond}    & \dslbnf & \dslnt{Perc} \dslalt \dslkw{not}\;\dsltm{c(}\,\dslnt{Perc}\,\dsltm{c)} \\
  \dslnt{Perc}    & \dslbnf & \dsltm{frontIsClear} \dslalt \dsltm{leftIsClear} \\
                  &         & \dslalt\;\dsltm{rightIsClear} \\
                  &         & \dslalt\;\dsltm{markersPresent} \\[1pt]
  \dslnt{Action}  & \dslbnf & \dsltm{move} \dslalt \dsltm{turnLeft} \dslalt \dsltm{turnRight} \\
                  &         & \dslalt\;\dsltm{pickMarker} \dslalt \dsltm{putMarker} \\
  \end{tabular}
  \end{tcolorbox}
  \caption{Karel DSL.}
  \label{fig:karel_dsl}
\end{subfigure}%
\hfill%
\begin{subfigure}[t]{0.333\linewidth}
  \centering
  \begin{tcolorbox}[
      enhanced, sharp corners, boxrule=0.5pt, colframe=black, colback=white,
      left=2pt, right=2pt, top=3pt, bottom=3pt,
      equal height group=dslboxes,
  ]
  \dslfont
  \setlength{\tabcolsep}{1pt}
  \renewcommand{\arraystretch}{1.1}
  \begin{tabular}{r@{\;}c@{\;}l}
  \dslnt{Prog}    & \dslbnf & \dslkw{DEF}\,\dslkw{run}\,\dsltm{m(}\;\dslnt{Stmt}\;\dsltm{m)} \\[1pt]
  \dslnt{Stmt}    & \dslbnf & \dslnt{Action} \dslalt \dslnt{Stmt}\;\dslnt{Stmt} \\
                  &         & \dslalt\;\dslnt{If} \dslalt \dslnt{IfElse} \dslalt \dslnt{While} \dslalt \dslnt{Repeat} \\[1pt]
  \dslnt{If}      & \dslbnf & \dslkw{IF}\;\dsltm{c(}\,\dslnt{Cond}\,\dsltm{c)}\;\dsltm{i(}\,\dslnt{Stmt}\,\dsltm{i)} \\
  \dslnt{IfElse}  & \dslbnf & \dslkw{IFELSE}\;\dsltm{c(}\,\dslnt{Cond}\,\dsltm{c)} \\
                  &         & \dsltm{i(}\,\dslnt{Stmt}\,\dsltm{i)}\;\dslkw{ELSE}\;\dsltm{e(}\,\dslnt{Stmt}\,\dsltm{e)} \\
  \dslnt{While}   & \dslbnf & \dslkw{WHILE}\;\dsltm{c(}\,\dslnt{Cond}\,\dsltm{c)}\;\dsltm{w(}\,\dslnt{Stmt}\,\dsltm{w)} \\
  \dslnt{Repeat}  & \dslbnf & \dslkw{REPEAT}\;\dslnt{N}\;\dsltm{r(}\,\dslnt{Stmt}\,\dsltm{r)} \\
                  &         & \quad{\tiny($\dslnt{N}\!\in\![0,19]$)} \\[1pt]
  \dslnt{Cond}    & \dslbnf & \dslnt{Perc} \dslalt \dslkw{not}\;\dsltm{c(}\,\dslnt{Perc}\,\dsltm{c)} \\
  \dslnt{Perc}    & \dslbnf & \dsltm{cursorAtStart} \\
                  &         & \dslalt\;\dsltm{cursorAtEnd} \dslalt \dsltm{listEmpty} \\
                  &         & \dslalt\;\dsltm{elemIsZero} \dslalt \dsltm{elemLtNext} \\[1pt]
  \dslnt{Action}  & \dslbnf & \dsltm{cursorLeft} \dslalt \dsltm{cursorRight} \\
                  &         & \dslalt\;\dsltm{pushZero} \dslalt \dsltm{pop} \dslalt \dsltm{swapNext} \\
                  &         & \dslalt\;\dsltm{incElem} \dslalt \dsltm{decElem} \\
  \end{tabular}
  \end{tcolorbox}
  \caption{List DSL.}
  \label{fig:list_dsl}
\end{subfigure}%
\hfill%
\begin{subfigure}[t]{0.333\linewidth}
  \centering
  \begin{tcolorbox}[
      enhanced, sharp corners, boxrule=0.5pt, colframe=black, colback=white,
      left=2pt, right=2pt, top=3pt, bottom=3pt,
      equal height group=dslboxes,
  ]
  \dslfont
  \setlength{\tabcolsep}{1pt}
  \renewcommand{\arraystretch}{1.1}
  \begin{tabular}{r@{\;}c@{\;}l}
  \dslnt{Prog}    & \dslbnf & \dslkw{DEF}\;\dslnt{Body}\;\dsltm{m)} \\[1pt]
  \dslnt{Body}    & \dslbnf & \dslnt{Op} \dslalt \dslkw{COMBINE}\;\dslnt{CF}\;\dslnt{OpCat}\;\dslnt{OpCat} \\[1pt]
  \dslnt{Op}      & \dslbnf & \dslkw{MAP}\;\dslnt{FN} \dslalt \dslkw{SHIFT}\;\dslnt{OFF} \dslalt \dslkw{HIST} \\
                  &         & \dslalt\;\dslkw{CMPC}\;\dslnt{CMP}\;\dslnt{C} \dslalt \dslkw{COUNT}\;\dslnt{CMP} \\
                  &         & \quad{\tiny(\dslkw{COUNT}: \dslnt{CMP}\,$\neq$\,\dsltm{EQ})} \\
  \dslnt{OpCat}   & \dslbnf & \dslkw{MAP}\;\dslnt{FN} \dslalt \dslkw{SHIFT}\;\dslnt{OFF} \\
                  &         & \dslalt\;\dslkw{CMPC}\;\dslnt{CMP}\;\dslnt{C} \dslalt \dslkw{TOK} \\[1pt]
  \dslnt{FN}      & \dslbnf & \dsltm{FN\_inc} \dslalt \dsltm{FN\_dec} \dslalt \dsltm{FN\_dbl} \\
                  &         & \dslalt\;\dsltm{FN\_mod2} \dslalt \dsltm{FN\_mod3} \\
  \dslnt{OFF}     & \dslbnf & \dsltm{OFF\_1} \dslalt \dsltm{OFF\_2} \dslalt \dsltm{OFF\_3} \\
                  &         & \dslalt\;\dsltm{OFF\_m1} \dslalt \dsltm{OFF\_m2} \dslalt \dsltm{OFF\_m3} \\
  \dslnt{CMP}     & \dslbnf & \dsltm{EQ} \dslalt \dsltm{LT} \dslalt \dsltm{LEQ} \\
                  &         & \dslalt\;\dsltm{GT} \dslalt \dsltm{GEQ} \dslalt \dsltm{NEQ} \\
  \dslnt{C}       & \dslbnf & \dsltm{C1} \dslalt \dsltm{C2} \dslalt \ensuremath{\cdots} \dslalt \dsltm{C18} \\
  \dslnt{CF}      & \dslbnf & \dsltm{CF\_add} \dslalt \dsltm{CF\_sub} \dslalt \dsltm{CF\_max} \\
                  &         & \dslalt\;\dsltm{CF\_min} \dslalt \dsltm{CF\_gt} \dslalt \dsltm{CF\_eq} \\
  \end{tabular}
  \end{tcolorbox}
  \caption{RASP DSL.}
  \label{fig:rasp_dsl}
\end{subfigure}
\caption{The Karel, List, and RASP domain-specific languages used in our experiments. Karel and List support classical control flow (sequencing, conditionals, bounded repetition, unbounded loops over a perception predicate) and differ in their state-dependent perceptions and primitive actions. A RASP program is a single sequence operation (\dslnt{Op}) or a \dslkw{COMBINE} of two categorical ones (\dslnt{OpCat}); \dslkw{DEF} and \dsltm{m)} are the start/end tokens.}
\label{fig:dsl}
\end{figure}

We illustrate the formulation with the Karel \citep{pattis1994karel} and List DSLs (Figure~\ref{fig:dsl}), which share an identical control-flow skeleton (sequencing, conditionals, bounded repetition, unbounded loops over a perception predicate) and differ in their state-dependent perceptions and primitive actions. RASP programs (Figure~\ref{fig:rasp_dsl}) have no intermediate states, so there $s_0 = (x, p)$ is an input $x$ and a position $p$, and $s_t = (x^{(t)}, \rho(x^{(t)}))$, with $x^{(t)}$ the input $x$ with its $p$-th token set to $t$.

Each state $s_{k,t}$ is mapped to a dense embedding $\mathbf{v}_{k,t} \in \mathbb{R}^d$ by a domain-specific encoder $\phi$. For grid-based domains such as Karel, $\phi$ is a Convolutional Neural Network applied to the visual grid; for symbolic domains such as list manipulation, $\phi$ is a small elementwise embedder; for RASP, it is a linear map followed by a residual MLP. In all cases, each trace $\tau_k$ is transformed into a sequence of latent vectors $V_k = (\mathbf{v}_{k,1}, \dots, \mathbf{v}_{k, T_k})$.

\subsection{Why signatures match programs}
\label{subsec:why_signatures}

Encoding each trace by its (log-)signature, rather than by a learned recurrent or attention-based summary, is the central design choice of our approach. Two foundational results from rough path theory ground this choice.

First, a program $\rho$ is determined, up to behavioral equivalence, by the distribution $\mu_\rho$ it induces over execution traces, with the randomness coming from the choice of initial state. \citet{chevyrev2016characteristic} show that the \emph{expected signature} $\mathbb{E}_{X \sim \mu_\rho}[S(X)]$ uniquely characterizes $\mu_\rho$ whenever it has an infinite radius of convergence, which holds automatically for bounded-length traces such as ours. Aggregating per-trace signatures across multiple executions of the same program therefore estimates an object that, in principle, suffices to identify the generating program.

Second, by Chen's identity \citep{chen1957integration}, the signature is multiplicative under any temporal split of a path:
\begin{equation}
    S(X_{[0, T]}) \;=\; S(X_{[0, t]}) \,\otimes\, S(X_{[t, T]}) \quad \text{for any } t \in [0, T].
    \label{eq:chen_compositional}
\end{equation}
This is an algebraic property of \emph{paths}, not of program syntax: traces whose embedded paths share a contiguous sub-segment (e.g., the same loop body iterated, the same inner subroutine called) share a corresponding signature factor, so structurally similar executions land in structurally similar regions of the embedding space. As such, signatures respect the temporal structure of any single execution, equipping the embedding manifold with closed-form algebraic structure under temporal sub-decomposition.

By contrast, none of the baseline encoders has a closed-form law relating the embedding of a trace to those of its parts: an LSTM's embedding depends on state order through a fixed recurrence, a Transformer encoder recovers order only because its positional encoding breaks the permutation equivariance of self-attention, and the truncated Fourier spectrum of a concatenation cannot be recovered from the spectra of its parts.

\subsection{Trace encoding via signature-based transforms}
Embedding a sequence given the embeddings of its elements (states in our case) typically involves a choice of an aggregation scheme, such as recurrence in LSTMs and pooling as well as [CLS] tokens for Transformers. In practice, this choice is nontrivial and depends on the task at hand. For example, recurrent aggregation famously suffers from forgetting and vanishing gradients in long sequences. The signature is a natural aggregator: it outputs a fixed-size embedding regardless of the sequence length, and its runtime and memory complexities are linear in the sequence length rather than quadratic as in Transformers. 

A more fundamental difference that sets the signature apart from LSTM and Transformer encoders is that it operates on geometry while the latter operate on content. The signature captures the geometric shape of the path traced out by the sequence of state embeddings, while LSTMs and Transformers learn a content-based summary. We argue that programmatic behavior is fundamentally geometric: the control flow of a program structures the execution into subroutines, loops, and conditionals that manifest as geometric features in the trace. By encoding geometry directly, the signature is naturally aligned with the underlying structure of program execution and bypasses content-level noise.

\paragraph{Positional augmentation.}
Standard path signatures satisfy \textit{tree-like equivalence}, meaning they are invariant to backtracking (e.g., a sequence of actions like \texttt{turnRight}, \texttt{turnLeft} that returns to the original state might be nullified). To resolve this and preserve the distinct execution history, we augment the state embeddings with positional encodings. We define a time-augmented path sequence $\tilde{V}_k$ where:
\begin{equation}
    \tilde{\mathbf{v}}_{k,t} = \mathbf{v}_{k,t} + \text{PE}(t)
\end{equation}
Here, $\text{PE}(t) \in \mathbb{R}^d$ is a fixed sinusoidal positional encoding. Because $\text{PE}(t)$ differs at every step, an action sequence that returns to an earlier state reaches a different point of the augmented path, so it no longer forms a backtracking segment that the signature ignores. In many standard signature applications, time is instead added as an extra channel, which rules out tree-like equivalence altogether. In our setting, we found that additive positional encodings work better on Karel and List domains. On RASP, where step $t$ of every trace sets the swept token to $t$, the log-signature encoder instead appends a one-hot encoding of $t$.

\paragraph{Log-signature encoder.}
\label{para:logsig_encoder}
We treat the sequence $\tilde{V}_k$ as a discrete approximation of a continuous path $X_k: [0, T_k] \to \mathbb{R}^d$. To handle variable-length traces within the batch, we pad $\tilde{\mathbf{v}}_{k}$ by repeating the embedding of the last state in the trace: $\tilde{\mathbf{v}}_{k,t} = \tilde{\mathbf{v}}_{k,T_k}, \forall t > T_k$. Each repeated point adds a zero increment, whose signature is the identity $\exp(0) = 1$, so by Chen's identity the padding leaves the signature unchanged; this is an instance of the invariance of the signature to reparameterization, as stalling at a point does not change the shape of the path. This is preferable to zero-padding, which introduces an artifact in the form of a false increment at the end of the path.

Before adding the positional encoding, we normalize every step of the embedded trace to unit length while keeping its direction: with $\Delta_{k,t} = \mathbf{v}_{k,t+1} - \mathbf{v}_{k,t}$, we set $\hat{\mathbf{v}}_{k,1} = \mathbf{v}_{k,1}$ and $\hat{\mathbf{v}}_{k,t+1} = \hat{\mathbf{v}}_{k,t} + \Delta_{k,t} / \sqrt{\lVert \Delta_{k,t} \rVert^2 + \epsilon^2}$, where the small constant $\epsilon$ keeps zero-length steps at zero, and use $\hat{\mathbf{v}}_{k,t}$ in place of $\mathbf{v}_{k,t}$ when forming $\tilde{\mathbf{v}}_{k,t}$. We apply this normalization because the signature has no weights of its own to rebalance steps of different sizes: its level-$\ell$ terms are sums of products of $\ell$ increments, so a few long steps dominate it. In List, for example, \texttt{pushZero} and \texttt{pop} shift several elements at once, and their steps in the embedded path are over ten times longer than those of \texttt{incElem} and \texttt{decElem}, so the in-place edits contribute little to the log-signature. With unit steps, every state change contributes equally, and the log-signature records which changes occurred and in what order rather than how far each one moved the embedding. 

We then compute the truncated log-signature of this padded path up to level $M=2$, denoted $\log(S^M(\tilde{V}_k))$. We choose $M=2$ as the smallest truncation level that retains pairwise interaction terms between coordinates of the embedded path; an ablation over $M \in \{1, 2, 3\}$ in Appendix~\ref{subsec:m_ablation} shows that going beyond $M=2$ yields no consistent benefit, at a substantial dimensional cost. The log-signature creates a fixed-size representation of the path regardless of the trace length $T_k$. Since the dimension of the log-signature $D_{\text{logsig}}(d, M)$ grows rapidly with the truncation level $M$, we project it back to a compact representation space via a learnable linear map $W_{\text{proj}} \in \mathbb{R}^{d \times D_{\text{logsig}}(d, M)}$ with bias $\mathbf{b}_{\text{proj}} \in \mathbb{R}^d$:
\begin{equation}
    \mathbf{h}_k = W_{\text{proj}} \cdot \log(S^M(\tilde{V}_k)) + \mathbf{b}_{\text{proj}} \in \mathbb{R}^d.
\end{equation}
The projection is the only trainable component of this encoder, so the underlying signature representation remains parameter-free.

\subsection{Program decoding}
To generate the program tokens, we employ a Transformer decoder. The decoder conditions on the set of trace embeddings $H = \{\mathbf{h}_1, \dots, \mathbf{h}_K\}$ through a cross-attention mechanism.
\begin{equation}
    P(\hat{\rho} | H) = \prod_{j=1}^{|\hat{\rho}|} P_{\theta}(y_j | y_{<j}, H)
\end{equation}
where $y_j$ is the $j$-th token of the predicted program $\hat{\rho}$. The decoder queries the trace memory $H$, allowing the model to attend to specific features of the execution traces (captured by the signatures) that correspond to specific program subroutines. Similar to \citet{bunel2018leveraging}, we mask out syntactically invalid tokens as they are easy to recognize. In doing so, we ensure that the decoder always outputs syntactically valid programs.

\subsection{Optimization}
The model is trained end-to-end by minimizing the cross-entropy loss between the predicted program distribution and the ground-truth program $\rho^* = (y^*_1, \dots, y^*_{L})$:
\begin{equation}
    \mathcal{L}(\theta) = - \sum_{j=1}^{L} \log P_{\theta}(y^*_j | y^*_{<j}, H)
\end{equation}

\paragraph{Contractivity and bootstrapping.}
\label{subsec:bootstrapping}
We empirically observe that our trace-to-program pipeline is naturally \emph{contractive}: in a sizable fraction of cases, the model produces a program $\hat\rho$ that is shorter than the ground-truth program $\rho^*$ but still reproduces every supervising trace. We exploit this property as a way to expose redundancy in the ground-truth labels without external supervision. Starting from the original training set $\mathcal{D}_0 = \{(\mathcal{T}_i, \rho^*_i)\}$, we interleave training with relabeling passes, one every five epochs after a five-epoch warm-up: at pass $t$, the current model generates a prediction $\hat\rho_i$ for each training instance, and we form $\mathcal{D}_{t+1}$ from $\mathcal{D}_t$ by replacing each label whenever the prediction is both \emph{valid} (executing $\hat\rho_i$ on each initial state in $\mathcal{T}_i$ reproduces the corresponding trace) and \emph{shorter} than the current label:
\begin{equation}
    \rho_i^{(t+1)} = \begin{cases}
        \hat\rho_i & \text{if } \hat\rho_i \text{ valid on } \mathcal{T}_i \text{ and } |\hat\rho_i| < |\rho_i^{(t)}|, \\
        \rho_i^{(t)} & \text{otherwise.}
    \end{cases}
\end{equation}
Training then continues on $\mathcal{D}_{t+1}$, and the total training-program length decreases monotonically. The cumulative length reduction and the fraction of training samples whose labels are ever shortened give a direct, encoder-derived lower bound on how much shorter the supervision could have been, with no external oracle required. We quantify this on redundancy-injected versions of our datasets (Section~\ref{subsec:datasets}) in Section~\ref{subsec:bootstrap_results}.

\section{Experiments}
\label{sec:experiments}

\subsection{Datasets}
\label{subsec:datasets}
We evaluate trace-to-program synthesis on three complementary program-induction domains: \emph{Karel}, a 2D grid-world programming language with visual state, \emph{List}, a symbolic list-manipulation language, and \emph{RASP}, a language for expressing Transformer computations as sequence operations. Together they cover a broad slice of the program-induction problem space: Karel is representative of visual and spatial-reasoning domains (robotics control, tile-based puzzles, agent navigation), List is representative of symbolic array- and string-manipulation domains (sorting, searching, counting, sequence transformations), and RASP is representative of functional sequence-to-sequence programs. For Karel and List we adopt a similar data generation process to that of \citet{trivedi2021learning}, with 100K training and 2500 test samples. To mitigate overfitting during training, we randomly sample 5 out of 6 traces for each program, and use all 6 traces during testing. An output program is marked as correct if it reproduces all input traces when executed on their respective initial states. 

\paragraph{RASP.} RASP \citep{weiss2021thinking} composes elementwise maps (\texttt{MAP}), positional shifts (\texttt{SHIFT}), thresholds (\texttt{CMPC}), and selector-width counts (\texttt{HIST}, \texttt{COUNT}) over a token sequence. RASP programs can be compiled into Transformer weights \citep{lindner2023tracr}, and Transformers can be trained so that they convert into human-readable programs \citep{friedman2023learning}. We use programs of nesting depth at most two (Figure~\ref{fig:rasp_dsl}) on length-$16$ sequences of tokens in $\{1, \dots, 18\}$, one per distinct non-constant behavior: $5{,}816$ for training and $2{,}500$ for testing, so every test program computes an unseen function. A trace is a \emph{sweep}: from a random input, the token at one position is set to $1, \dots, 18$ in turn, and each step is the pair of input and output sequences, a point in $\mathbb{R}^{32}$. Each program has $K{=}8$ sweeps at distinct positions, of which we sample $4$ during training and use all $8$ for testing; redrawing the sweeps of the training programs gives $32$K training samples. A decoded program is correct if it reproduces every output of every sweep.

\paragraph{Redundancy-injected datasets.} Bootstrapping (Section~\ref{subsec:bootstrapping}) can only shorten a label if a shorter program reproduces the same traces, and our datasets contain little such redundancy: on the Karel and List test splits, a greedy search over trace-preserving local edits shortens about a quarter of the programs but removes only $4.5\%$ and $5.2\%$ of their tokens. To highlight the contractive effect of our pipeline, we therefore build redundancy-injected versions of the three datasets, denoted Karel$^{+}$, List$^{+}$, and RASP$^{+}$, by rewriting the ground-truth programs. In each split, half of the programs are rewritten with extra structure that leaves their actions (Karel, List) or outputs (RASP) unchanged on every input, so injected tokens make up about $20\%$ of the total. For Karel and List, the rewrites invert standard simplification rules, e.g.\ $S \to$ \dslkw{REPEAT} \texttt{1} $(S)$ or $S \to$ \dslkw{IFELSE} $c$ $(S)$ \dslkw{ELSE} $(S)$. For RASP, we rewrite one categorical operation into an equivalent \dslkw{COMBINE} of two, e.g.\ \texttt{CMPC EQ $c$} into \dslkw{COMBINE} \texttt{min (CMPC GEQ $c$) (CMPC LEQ $c$)}, allowing \dslkw{COMBINE}s to nest. The traces are unchanged, and we use these datasets only for the bootstrapping experiment (Finding~\#2).

\subsection{Baselines}
We compare our log-signature trace encoder against three classical baselines that differ from it only in the trace encoder, up to one minior exception for Fourier (Appendix~\ref{subsec:optimization_details}).

\paragraph{LSTM.} We pass the state embeddings as the input sequence to an LSTM \citep{hochreiter1997long}, and the last hidden state is taken as the trace embedding. Although the temporal order is implicit in LSTM recurrence, we found that adding sinusoidal positional encoding to the input state sequence improved its performance.

\paragraph{Transformer.} We use a 2-layer Transformer encoder \citep{vaswani2017attention} with 8 self-attention heads and sinusoidal positional encoding added to the input sequence. We found that adding more layers did not significantly improve performance, and resulted in a collapse in the case of 8 or more layers. We obtain the trace embeddings by mean-pooling over the output state embeddings.

\paragraph{Fourier transform.} We compute the truncated orthonormal real-to-complex DFT \citep{cooley1965algorithm} of the (positionally augmented) state sequence, retaining the $m{=}9$ lowest-frequency modes and stacking their real and imaginary parts into the trace embedding. Full equations are given in Appendix~\ref{subsec:fourier_details}.

\subsection{Hyperparameters} All models are trained for 100 epochs with a batch size of 32 and an embedding size of 256 ($d{=}32$ for the RASP log-signature encoder). We use the Adam optimizer \citep{kingma2014adam} with a learning rate of $10^{-4}$ (details in Appendix~\ref{subsec:optimization_details}). We clip the global gradient norm at 0.5 \citep{pascanu2013difficulty}. The training pipeline is implemented using \textsc{PyTorch}, including a \textsc{PyTorch}-native reimplementation of \textsc{signatory} \citep{kidger2021signatory} for differentiable log-signature computations, and runs on a single NVIDIA A100 GPU.

\subsection{Results}
We organize our findings around three claims, each stated up front and then supported by the corresponding evidence. We report \emph{semantic accuracy}: the fraction of test programs whose decoded program is correct in the sense of Section~\ref{subsec:datasets}, i.e., it reproduces every given trace (Karel, List) or output (RASP), whether or not it matches the ground-truth program token for token.

\begin{figure}[h]
  \centering
  \resizebox{\linewidth}{!}{\input{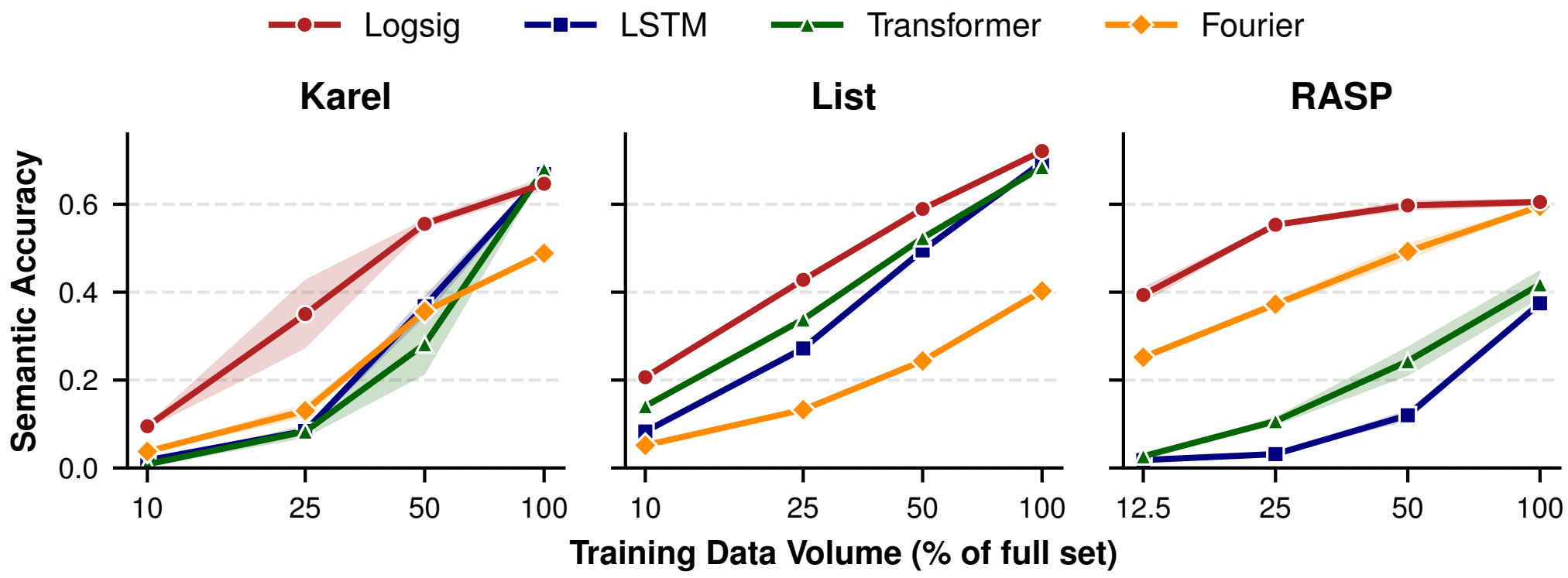}}
  \caption{Generalization performance across training data volumes for each trace encoder on the Karel, List, and RASP domains. Training data volume is a percentage of each domain's full training set, on a log scale. Each point is the highest test semantic accuracy over the 100 training epochs, averaged over 3 random seeds. The test set (2500 samples) is the same for all volumes.}
  \label{fig:data_eff_perf}
\end{figure}

\paragraph{Finding \#1: The log-signature encoder outperforms the LSTM, Transformer, and Fourier baselines in almost every setting, with its largest margins when training data is scarce.}
\label{subsec:finding_main_perf}

Figure~\ref{fig:data_eff_perf} reports test semantic accuracy across training data volumes for each encoder on the Karel, List, and RASP domains. The log-signature encoder is the most accurate encoder at 11 of the 12 (domain, volume) settings. The exception is Karel at full volume, where the Transformer and the LSTM lead it by more than the spread across seeds. The advantage is largest in the low-data regime: with a quarter of the training data, the log-signature encoder nearly triples the accuracy of the strongest baseline on Karel and clearly leads it on List and RASP, where the margin narrows at full volume. The parametric encoders need the most data: with a quarter of the training data, the LSTM and Transformer barely learn Karel, and on RASP they stay far behind the log-signature encoder at every volume. The Fourier encoder beats the LSTM and Transformer at low volumes on Karel and at every volume on RASP, but trails them on Karel at full volume and is the weakest encoder on List at every volume. We further find that the log-signature encoder reaches $90\%$ of its best accuracy in fewer epochs than the LSTM and Transformer at every volume where these learn, and the Fourier encoder does so in most settings. Two ablations indicate that the advantage of the log-signature encoder does not stem from the size of its representation. When we vary the width $d$ of the state embeddings at $50\%$ of the training data volume, even the narrowest log-signature encoder is more accurate than every baseline at any width on List and RASP, and on Karel it leads from $d{=}64$ onwards (Figure~\ref{fig:width_sweep}, Appendix~\ref{subsec:width_ablation}). Furthermore, the log-signature encoder performs nearly identically to its full-signature counterpart on all three domains, while using half as many features and, on Karel and List, about a third fewer parameters (Figure~\ref{fig:sig_vs_logsig}, Appendix~\ref{subsec:sig_vs_logsig}).

\begin{figure*}[t]
    \centering
    \resizebox{\textwidth}{!}{\input{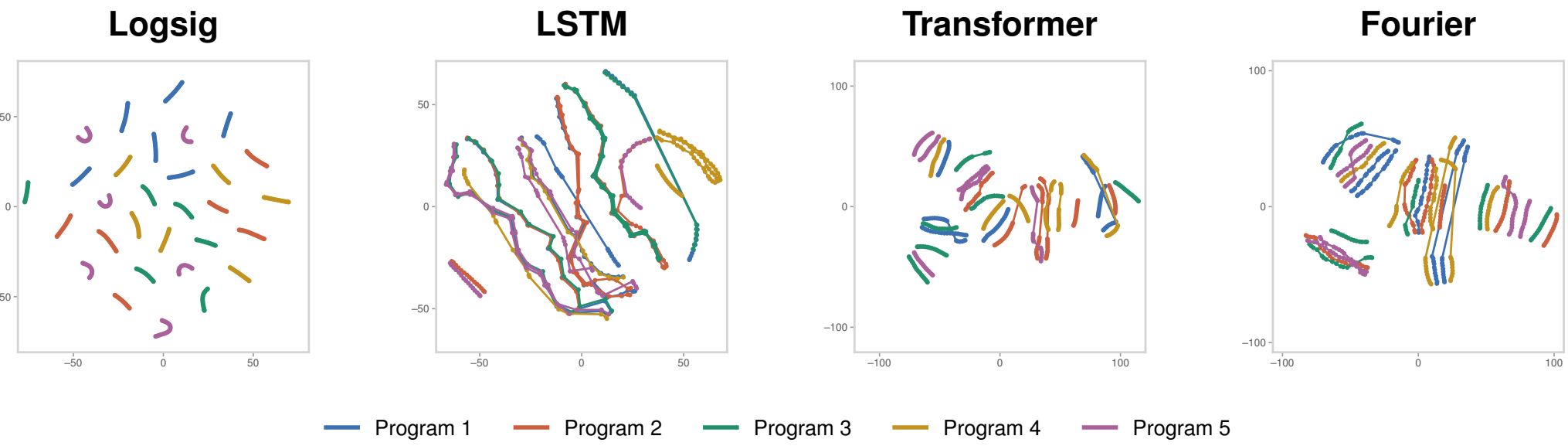}}
    \caption{t-SNE projections of the embedded traces that each trace encoder reads on Karel, for five test programs with six traces each. Each line is one trace, its points are the trace's states in temporal order, and the traces of a program share a color. For a common view, the state embeddings of all four encoders are normalized to unit-length steps as in Section~\ref{para:logsig_encoder} before the positional encoding is added. Each panel is a separate t-SNE fit.}
    \label{fig:path_tsne}
\end{figure*}

\paragraph{Finding \#2: Bootstrapping exposes substantial redundancy in the ground-truth supervision, with no external oracle required.}
\label{subsec:bootstrap_results}

We run the bootstrapping procedure of Section~\ref{subsec:bootstrapping} on each (encoder, task) configuration, using the redundancy-injected datasets Karel$^{+}$, List$^{+}$, and RASP$^{+}$ (Section~\ref{subsec:datasets}), and report two quantities: \emph{Tokens removed}, the cumulative percent reduction in total training-program token count from the labels of the redundancy-injected dataset to the final labels; and \emph{Programs shortened}, the percent of training programs whose label was shortened at least once during the procedure. We frame this as a \emph{measurement} rather than a self-improvement claim: the procedure replaces each ground-truth label only when the model has produced a strictly shorter program that is verified to reproduce every supervising trace, so both quantities are encoder-derived lower bounds on how compressible the supervision was. Stronger encoders expose more of this redundancy because they recover more valid alternatives; the absolute numbers therefore vary across the rows of Table~\ref{table:bootstrap}, but every cell is a lower bound on the same underlying quantity. Notably, every encoder, including the parametric LSTM and Transformer baselines and the spectral Fourier baseline, produces nontrivial values of both quantities, so contractivity is better understood as a property of the trace-to-program \emph{pipeline} than as a property unique to any one encoder.

\begin{table}[h]
  \caption{Reduction statistics in the ground-truth supervision of the redundancy-injected datasets (Section~\ref{subsec:datasets}), uncovered by bootstrapping.}
  \vspace{1em}
  \label{table:bootstrap}
  \centering
  \footnotesize
  \setlength{\tabcolsep}{3.5pt}
  \begin{tabular}{l cc cc cc}
    \toprule
    & \multicolumn{2}{c}{\textbf{\boldmath Karel$^{+}$}} & \multicolumn{2}{c}{\textbf{\boldmath List$^{+}$}} & \multicolumn{2}{c}{\textbf{\boldmath RASP$^{+}$}} \\
    \cmidrule(lr){2-3} \cmidrule(lr){4-5} \cmidrule(lr){6-7}
    \textbf{Encoder} & Tokens & Programs & Tokens & Programs & Tokens & Programs \\
    & removed (\%) & shortened (\%) & removed (\%) & shortened (\%) & removed (\%) & shortened (\%) \\
    \midrule
    LSTM          & \textbf{21.11} & \textbf{58.66} & 22.37 & 59.42 & 13.41 & 32.33 \\
    Transformer   & 20.92 & 57.81 & 22.58 & 59.82 & 18.26 & 44.08 \\
    Fourier       & 18.43 & 52.89 & 18.19 & 49.62 & 18.64 & 44.98 \\
    Log-Signature & 21.07 & 58.59 & \textbf{22.70} & \textbf{59.88} & \textbf{19.84} & \textbf{47.92} \\
    \bottomrule
  \end{tabular}
\end{table}

\paragraph{Finding \#3: The log-signature encoder induces a more interpretable latent geometry.}
\label{subsec:finding_geometry}

We investigate the geometry of the learned representations using t-SNE \citep{maaten2008visualizing}. Figure~\ref{fig:path_tsne} shows the projection of high-dimensional latent state embeddings (the state embeddings each trace encoder reads, after the unit-step normalization and positional encoding) into a reduced two-dimensional space. This is essentially a visualization of our main thesis: that programs have an inherent shape that manifests in their traces. As shown in Figure~\ref{fig:path_tsne}, the embeddings of the log-signature-based model are much more spatially structured and interpretable: states from the same execution trace remain clustered and form separate chains of nearby points within the embedding manifold, without crossing into each other as is the case with the baselines. This geometry reveals that the model preserves the sequential nature of the data. Figure~\ref{fig:path_tsne_full} in Appendix~\ref{subsec:state_geometry_all} extends this view to the List and RASP domains.

\section{Conclusion}
\label{sec:conclusion}

\paragraph{Summary.}  In this paper, we brought a rough-path-theoretic perspective to program induction, exploring the path log-signature transform as a trace encoder for trace-to-program synthesis. Our results have shown an inherent advantage of the log-signature encoder over traditional sequence models. However, this is not to say the signature transform should necessarily be treated as a complete replacement for other sequence encoders, but rather to establish grounds for future research in employing geometric methods for program learning. The essence of this work is demonstrating that geometry potentially reveals more about programmatic structure than content. Furthermore, we have empirically shown that the trace-to-program pipeline is naturally \emph{contractive}, which, as our experiments on redundancy-injected datasets show, serves as a convenient tool for exposing redundancy without any external oracle. Finally, we have shown that the log-signature encoder induces a markedly more interpretable latent geometry than classical sequence models.

\paragraph{Limitations.} The truncated log-signature provides a compact representation for moderately complex traces, but the dimensionality of the feature space $D_{\text{logsig}}$ still grows exponentially with the truncation level $M$. We keep $M$ small in practice (Appendix~\ref{subsec:m_ablation}), but very-long-horizon programs that require fine-grained higher-order interactions across the full trace remain a challenging case for the approach.

\begin{ack}
This work was partially supported by the European Union with ERC Grant HYPER (No. 101055412).
Views and opinions expressed are however those of the authors only and do not necessarily reflect
those of the European Union or the European Research Council Executive Agency. Neither the
European Union nor the granting authority can be held responsible for them.
\end{ack}

\bibliographystyle{plainnat}
\bibliography{bibliography}

\clearpage
\appendix

\section{Appendix}
\label{app:additional_experiments}

This appendix collects probe experiments that complement the three main findings in Section~\ref{sec:experiments}. Each subsection is self-contained and may be read independently of the others.

\subsection{Signature truncation depth ablation}
\label{subsec:m_ablation}

\begin{figure}[h]
  \centering
  \resizebox{\linewidth}{!}{\input{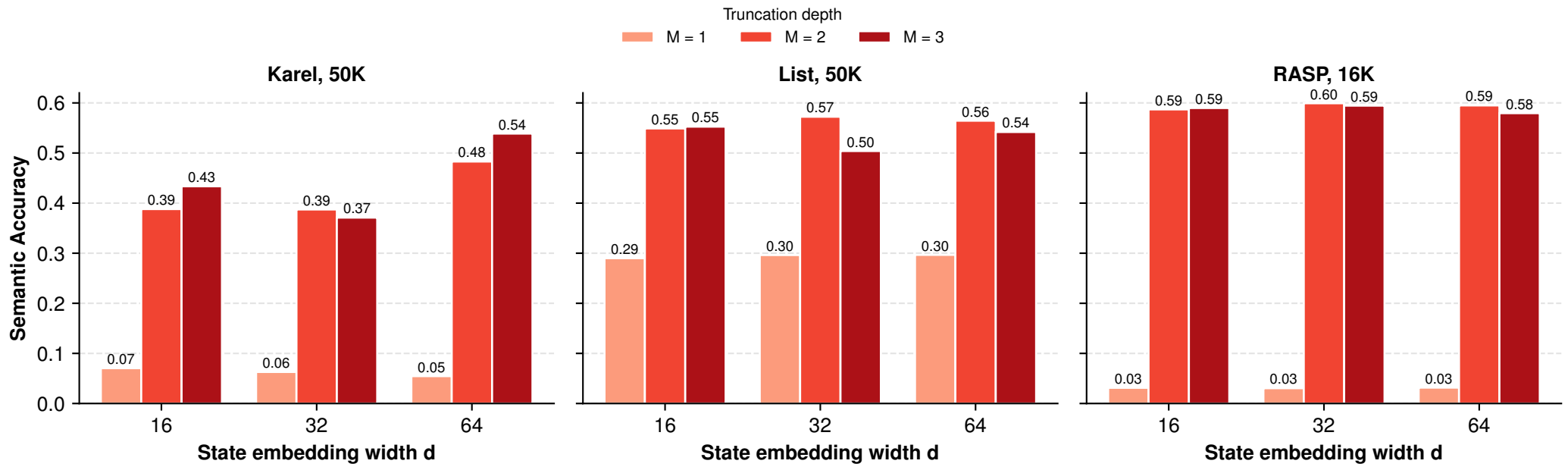}}
  \caption{Truncation depth ablation at half the training data volume (50K samples for Karel and List, 16K for RASP). Semantic accuracy (the highest over the 100 training epochs) on the 2500-sample test set as a function of the log-signature truncation level $M$ and the underlying state embedding dimension $d$.}
  \label{fig:m_sweep}
\end{figure}

The truncation level $M$ controls the maximum interaction order captured by the log-signature: $M{=}1$ retains only the displacement of the path, $M{=}2$ adds pairwise interactions (L\'evy areas), and higher $M$ captures progressively higher-order shape statistics at exponential dimensional cost. We jointly sweep $M \in \{1, 2, 3\}$ and the underlying state embedding dimension $d \in \{16, 32, 64\}$ at half of each domain's training data volume, and report semantic accuracy on the standard 2500-sample test set. Figure~\ref{fig:m_sweep} shows that $M{=}1$ falls far short of $M{=}2$ on every domain and dimension, and stays near zero on Karel and RASP, confirming that pairwise interactions are the load-bearing component of the representation. Going further to $M{=}3$ does not help on List and RASP, where it matches $M{=}2$ at $d{=}16$ and is at or below it at larger $d$. On Karel, $M{=}3$ improves on $M{=}2$ at $d{=}16$ and $d{=}64$, but not at $d{=}32$. Since $M{=}3$ multiplies the number of log-signature features by more than $40$ at $d{=}64$ for a gain on one domain only, we use $M{=}2$ in the main experiments.

\subsection{Signature versus log-signature}
\label{subsec:sig_vs_logsig}

\begin{figure}[h]
  \centering
  \resizebox{\linewidth}{!}{\input{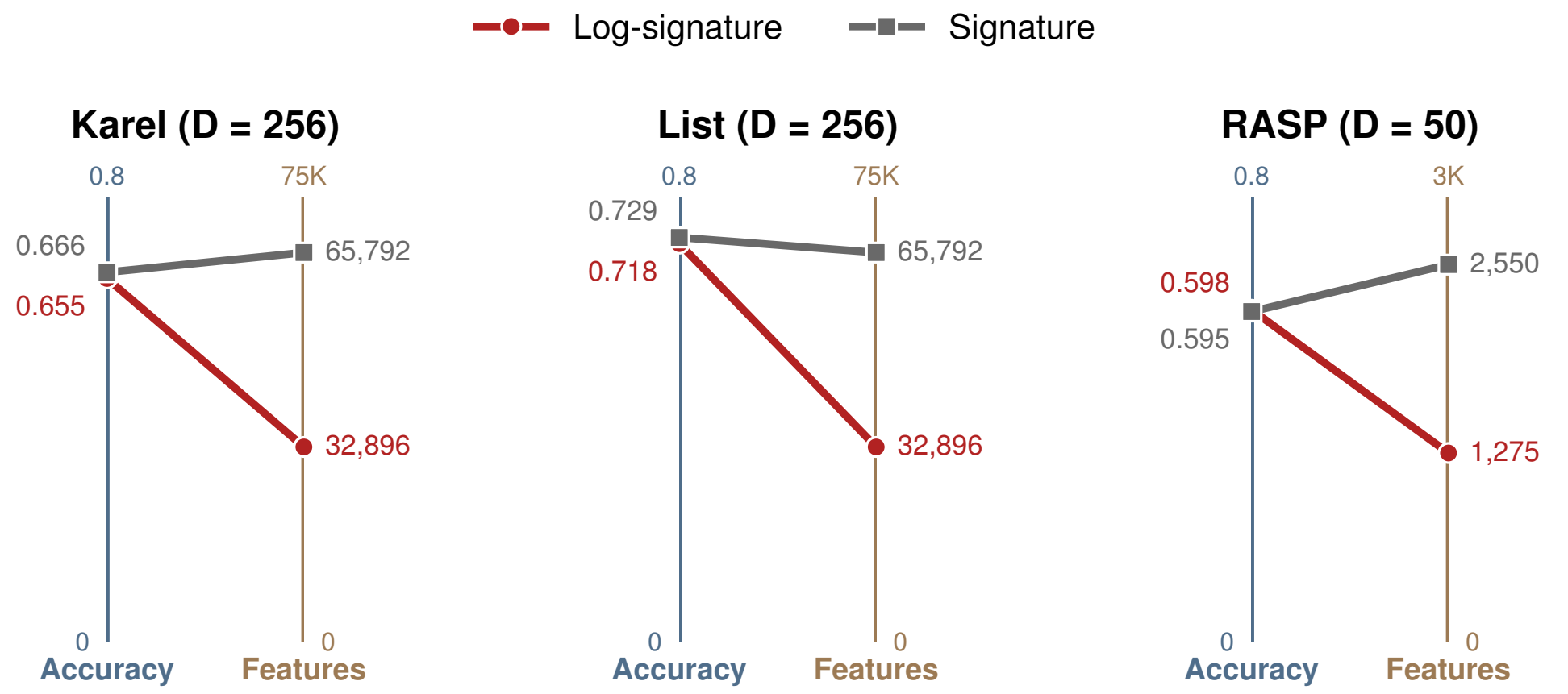}}
  \caption{Signature versus log-signature encoder at truncation level $M{=}2$, trained on each domain's full training set.}
  \label{fig:sig_vs_logsig}
\end{figure}

The log-signature is the compact alternative to the full signature: it removes the redundant terms generated by the shuffle product. At $M{=}2$ these redundant terms make up half of the signature. For a path with $D$ channels, the signature has $D + D^2 = D(D{+}1)$ non-constant coordinates and the log-signature $D(D{+}1)/2$. The extra coordinates are the symmetric part of the second level, which by the shuffle identity equals $\tfrac{1}{2}\,\mathbf{X}^{(1)}_{0,T} \otimes \mathbf{X}^{(1)}_{0,T}$, i.e.\ products of first-level coordinates, while its antisymmetric part is the L\'evy area that the log-signature keeps. To test whether these redundant coordinates help the learned projection, we replace the log-signature with the signature at $M{=}2$, keep every other setting of the main experiments, and train on each domain's full training set (100K samples for Karel and List, 32K for RASP). Figure~\ref{fig:sig_vs_logsig} shows that doubling the number of features leaves semantic accuracy essentially unchanged: the two encoders differ by at most $0.011$ on any domain, in either direction. Since the input of the projection doubles, the signature model is also larger, with about $50\%$ more parameters on Karel and List. The log-signature thus matches the signature at half the representation size, which is why we use it throughout.

\subsection{State embedding width}
\label{subsec:width_ablation}

\begin{figure}[h]
  \centering
  \resizebox{\linewidth}{!}{\input{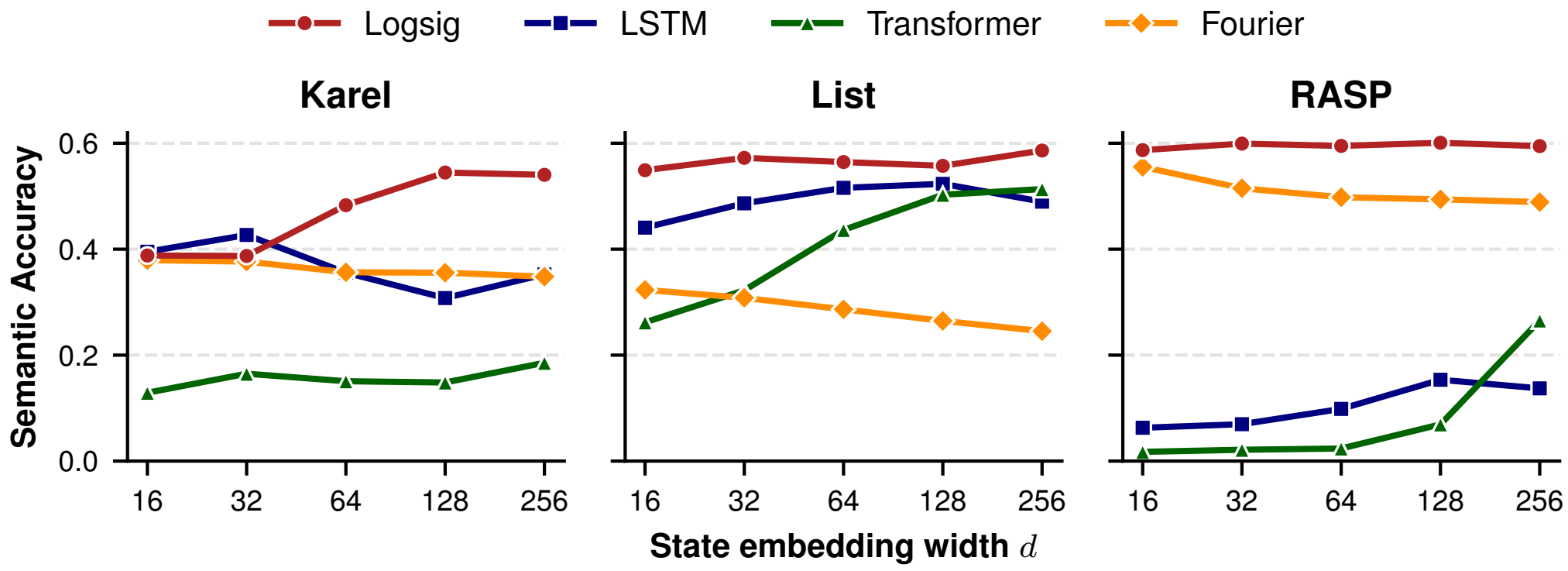}}
  \caption{Semantic accuracy of each trace encoder as a function of the state embedding width $d$, at $50\%$ of each domain's training data. Each point is the highest test semantic accuracy over the 100 training epochs on the 2500-sample test set.}
  \label{fig:width_sweep}
\end{figure}

At $M{=}2$, the log-signature of a $d$-dimensional path has $\mathcal{O}(d^2)$ coordinates, so the advantage of the log-signature encoder in Finding~\#1 could come from this large representation rather than from the structure of the signature. To test this, we vary the width $d \in \{16, 32, 64, 128, 256\}$ of the state embeddings fed to every trace encoder, at $50\%$ of each domain's training data volume and with all other settings as in the main experiments. For the LSTM only the input width changes, the Transformer operates at width $d$ throughout, and the number of Fourier features grows linearly with $d$. Figure~\ref{fig:width_sweep} shows that the log-signature encoder is the least sensitive to this choice. On List and RASP its accuracy is nearly flat across the whole range, and even its narrowest variant is more accurate than every baseline at any width. Karel is the exception: there the log-signature encoder gains substantially from wider embeddings, and the LSTM matches or exceeds it at the two smallest widths. The baselines respond to width in different ways. The Transformer is the most width-hungry encoder: it improves markedly as $d$ grows on List and RASP but trails far behind at small widths, and it stays low at every width on Karel. The Fourier encoder shows the opposite trend, being most accurate at the smallest width and degrading as $d$ grows on all three domains. The LSTM peaks at an intermediate width and remains far behind on RASP throughout. Overall, the ranking of the main experiments does not hinge on the chosen width: the log-signature encoder leads at every width on List and RASP and from $d{=}64$ onwards on Karel, and on two of the three domains a narrow state embedding, and hence a small log-signature, suffices.

\subsection{State-level latent geometry on all domains}
\label{subsec:state_geometry_all}

\begin{figure}[h]
  \centering
  \resizebox{\linewidth}{!}{\input{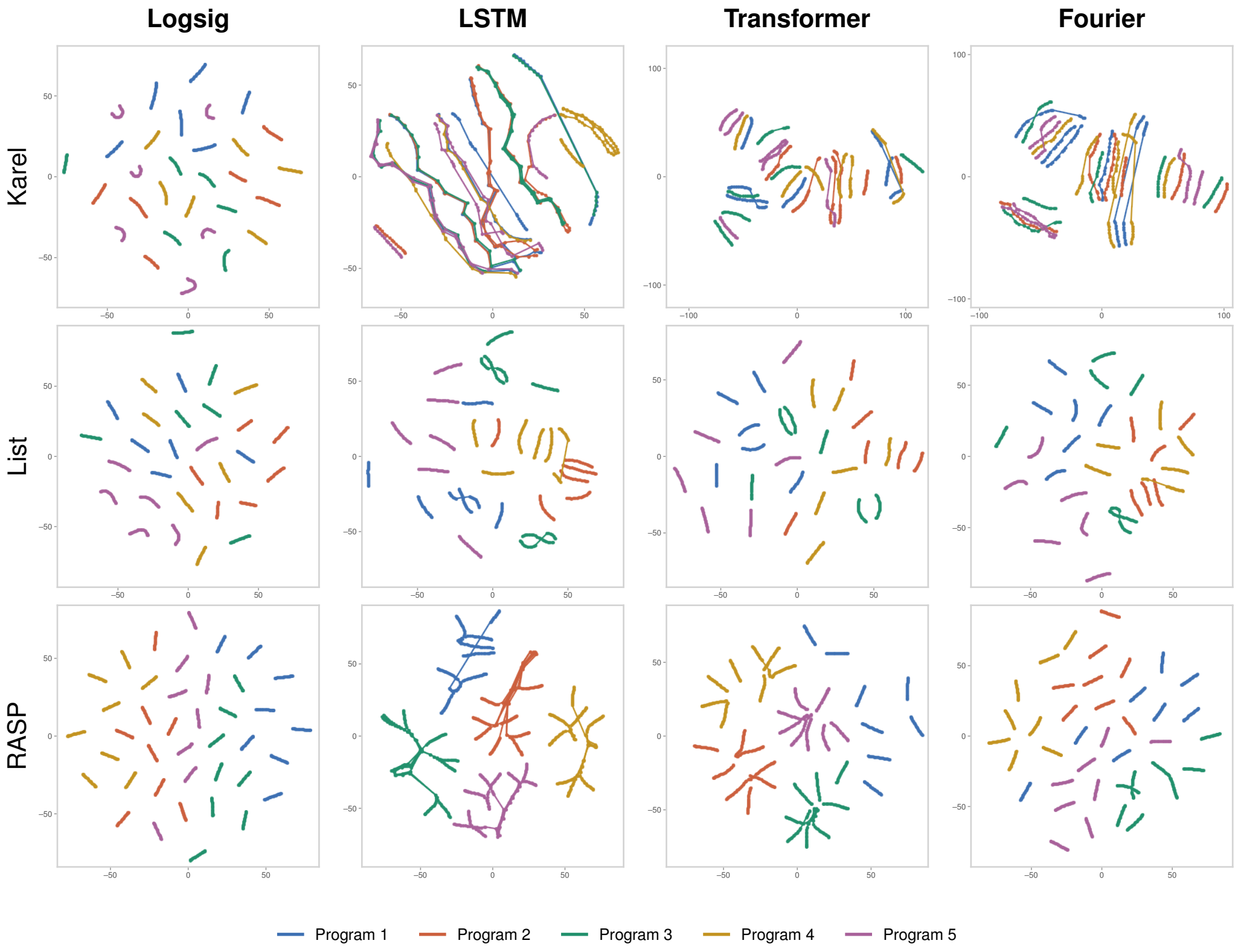}}
  \caption{State-level t-SNE projections as in Figure~\ref{fig:path_tsne}, for each trace encoder (columns) on the Karel, List, and RASP domains (rows), five test programs per domain with six traces each on Karel and List and eight on RASP. Each line is one trace, the traces of a program share a color, and each panel is a separate t-SNE fit.}
  \label{fig:path_tsne_full}
\end{figure}

Figure~\ref{fig:path_tsne_full} extends the state-level view of Finding~\#3 (Figure~\ref{fig:path_tsne}) from Karel to the List and RASP domains, with the same projection and the same unit-step normalization of every encoder's state embeddings. To quantify how well the traces stay apart, we take, for every state, its five nearest states in the full embedding space (over 20 test programs, before the t-SNE projection) and count the share that belongs to the same trace. The log-signature encoder scores $1.00$ on all three domains: the nearest neighbors of every state lie on its own trace. On Karel, all three baselines fall far short of this, with most of a state's neighbors coming from other traces. On RASP, they range from about half for the LSTM to nearly perfect for the Fourier encoder, and on List all four encoders keep most traces apart.

\subsection{Encoder forward-pass wall-clock scaling}
\label{subsec:wall_clock}

\begin{figure}[h]
  \centering
  \resizebox{0.6\textwidth}{!}{\input{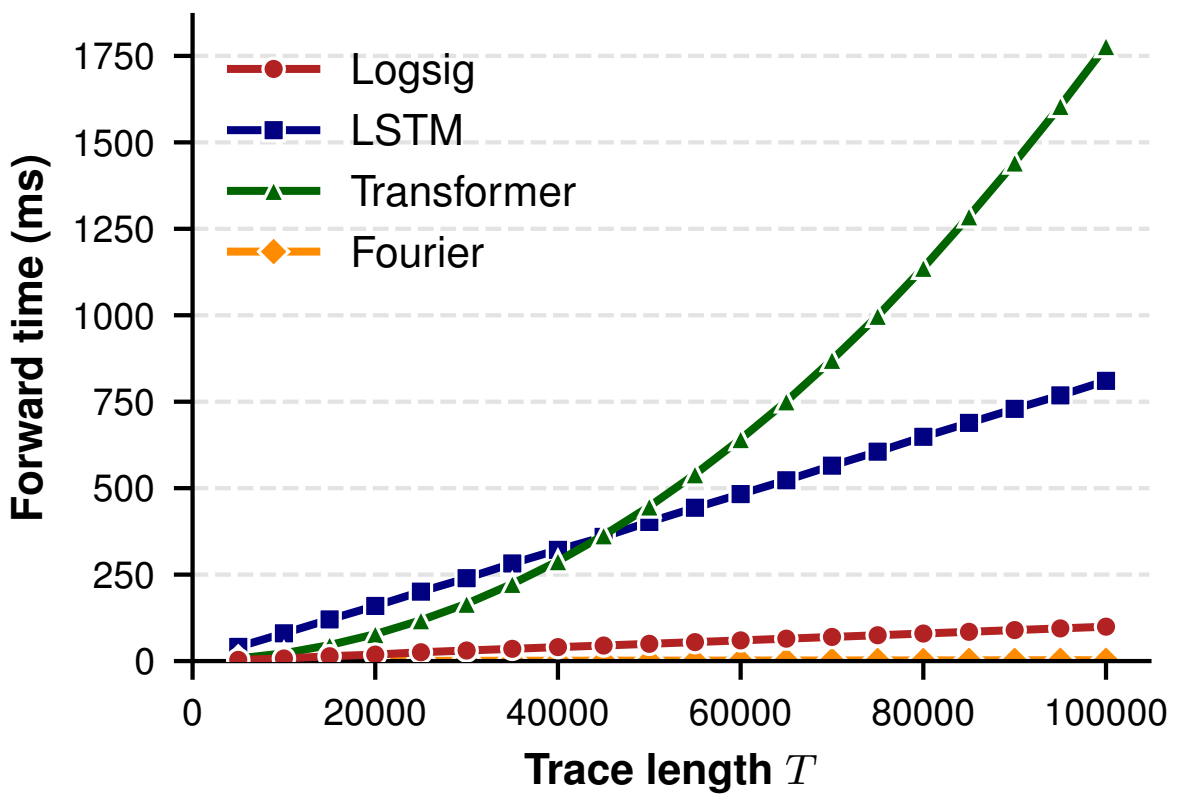}}
  \caption{Forward-pass wall-clock time of each trace encoder as a function of the trace length $T$, on one trace of width-$256$ state embeddings on one NVIDIA A100 GPU, with each encoder configured as in the Karel and List experiments. Only the trace encoder is timed. Each point is the mean of $5$ timed passes after $3$ warm-up passes; the standard deviations are below $2$\,ms.}
  \label{fig:encoder_runtime}
\end{figure}

To check the linear-time claim empirically, we time the forward pass of each trace encoder on a single trace of length $T$, up to $T{=}10^5$, on one NVIDIA A100 GPU. Each encoder has its configuration from the Karel and List experiments, with state embeddings of width $256$, and only the trace encoder itself is timed, not the state encoder, the set encoder (Appendix~\ref{subsec:optimization_details}) or the program decoder around it. Figure~\ref{fig:encoder_runtime} shows that the log-signature and LSTM encoders grow linearly with $T$, while the Transformer grows quadratically, as its attention compares every pair of steps. The Transformer overtakes the LSTM at around $45{,}000$ steps, and at $T{=}10^5$ it is about $18$ times slower than the log-signature encoder, which is in turn about $8$ times faster than the LSTM. The Fourier encoder, whose FFT is cheap at these lengths, is the fastest throughout.

\subsection{Fourier encoder: full definition}
\label{subsec:fourier_details}

For a single trace, we drop the trace index $k$: let $\mathbf{v}_1, \dots, \mathbf{v}_T \in \mathbb{R}^d$ be its state embeddings and $\mathbf{u}_t = \mathbf{v}_t + \text{PE}(t)$ their positionally augmented version, with the additive sinusoidal positional encoding $\text{PE}$ of Section~\ref{para:logsig_encoder}. The Fourier encoder applies the orthonormal DFT to each of the $d$ coordinates of $(\mathbf{u}_t)_{t=1}^{T}$, keeps the $m = 9$ lowest-frequency modes, and stacks their real and imaginary parts into the trace embedding $\mathbf{Y} \in \mathbb{R}^{m \times 2d}$:
\begin{align*}
    \hat{\mathbf{u}}_f &= \frac{1}{\sqrt{T}} \sum_{t=1}^{T} \mathbf{u}_t \, e^{-2\pi \mathrm{i} f (t-1)/T} \in \mathbb{C}^d, \qquad f = 0, \dots, m-1, \\
    \mathbf{y}_f &=
    \begin{bmatrix}
        \operatorname{Re}(\hat{\mathbf{u}}_f) \\
        \operatorname{Im}(\hat{\mathbf{u}}_f)
    \end{bmatrix}
    \in \mathbb{R}^{2d}, \\
    \mathbf{Y} &= \big[ \mathbf{y}_0, \mathbf{y}_1, \dots, \mathbf{y}_{m-1} \big]^\top.
\end{align*}

\subsection{Further optimization details}
\label{subsec:optimization_details}

\paragraph{Set encoder.} Before decoding, the $K$ trace embeddings $\mathbf{h}_1, \dots, \mathbf{h}_K$ of a program pass through a set encoder, a 2-layer Transformer encoder with 8 heads and no positional encoding, which lets each trace embedding attend to the others. We treat it as part of the program decoder, whose cross-attention reads the $K$ outputs of the set encoder rather than $H$ itself. The Fourier encoder is the one exception and is trained without the set encoder on all three domains, since with it the Fourier model did not learn on Karel at any training data volume.

\paragraph{Weight decay.} All models are trained with Adam and without weight decay, except the LSTM on RASP, which we train with AdamW \citep{loshchilov2019decoupled} and a weight decay of $1.0$ on its weight matrices. Without weight decay, the RASP LSTM does not learn: its semantic accuracy stays at about $0.03$ or below at every state embedding width and training data volume we tried, and weight decay rescues it partly.

\subsection{Compute resources}
\label{subsec:compute}

Every run uses a single NVIDIA A100 GPU. At half the training data volume, a training run takes about 4\,h on Karel, 1--2\,h on List, and 0.5\,h on RASP, and the time grows roughly in proportion to the volume. A bootstrapping run takes about 10\,h on Karel$^{+}$, 5\,h on List$^{+}$ (with two runs sharing one GPU), and 1--2\,h on RASP$^{+}$. The experiments reported in this paper total roughly 500 GPU-hours; the full project, including preliminary and discarded configurations, used more.

\end{document}